\documentclass[10pt,twocolumn,letterpaper]{article}

\usepackage{cvpr}              

\usepackage{booktabs}  
\usepackage{multirow}  
\usepackage{amsthm}  
\usepackage{nicefrac}  
\usepackage{makecell}  
\usepackage{pgfplots}  
\usepackage{rotating}  
\usepackage{tikz}  
\usetikzlibrary{shapes.geometric, arrows}
\usepackage{varwidth}  
\usepackage[export]{adjustbox}  
\usepackage{tabularray}  
    \UseTblrLibrary{varwidth}  
\usepackage{subcaption}  

\newtheorem{example}{Example}

\usepackage{amsmath,amsfonts,bm}

\usepackage{amsthm}
\usepackage{mathtools}

\newcommand{\lsbb}[2]{
    \ifthenelse{\equal{#2}{0}}
    {\it{BL}_{#1, #2}}
    {\it{BL}_{#1, #2}}
}
\newcommand{\lsbbcoeff}[2]{
    \ifthenelse{\equal{#2}{0}}
    {C_{\it{BL}_{#1, #2}}}
    {C_{\it{BL}_{#1, #2}}}
}
\newcommand{\lsbbconst}[2]{
    \ifthenelse{\equal{#2}{0}}
    {c_{\it{BL}_{#1, #2}}}
    {c_{\it{BL}_{#1, #2}}}
}
\newcommand{\rlsbb}[2]{
    \ifthenelse{\equal{#2}{0}}
    {\overline{BL}_{#1, #2}}
    {\overline{BL}_{#1, #2}}
}
\newcommand{\rlsbbcoeff}[2]{
    \ifthenelse{\equal{#2}{0}}
    {\overline{C}_{\it{BL}_{#1, #2}}}
    {\overline{C}_{\it{BL}_{#1, #2}}}
}
\newcommand{\rlsbbconst}[2]{
    \ifthenelse{\equal{#2}{0}}
    {\overline{c}_{\it{BL}_{#1, #2}}}
    {\overline{c}_{\it{BL}_{#1, #2}}}
}

\newcommand{\usbb}[2]{
    \ifthenelse{\equal{#2}{0}}
    {\it{BU}_{#1, #2}}
    {\it{BU}_{#1, #2}}
}
\newcommand{\usbbcoeff}[2]{
    \ifthenelse{\equal{#2}{0}}
    {C_{\it{BU}_{#1, #2}}}
    {C_{\it{BU}_{#1, #2}}}
}
\newcommand{\usbbconst}[2]{
    \ifthenelse{\equal{#2}{0}}
    {c_{\it{BU}_{#1, #2}}}
    {c_{\it{BU}_{#1, #2}}}
}

\newcommand{\loss}{\mathcal{L}}

\def\eqref#1{equation~\ref{#1}}

\def\1{\bm{1}}

\def\eps{{\epsilon}}

\def\va{{\bm{a}}}
\def\vb{{\bm{b}}}

\def\vl{{\bm{l}}}

\def\vu{{\bm{u}}}

\def\vx{{\bm{x}}}
\def\vy{{\bm{y}}}
\def\vz{{\bm{z}}}

\def\mA{{\bm{A}}}
\def\mB{{\bm{B}}}
\def\mC{{\bm{C}}}
\def\mD{{\bm{D}}}

\def\mG{{\bm{G}}}

\def\mI{{\bm{I}}}
\def\mJ{{\bm{J}}}
\def\mK{{\bm{K}}}
\def\mL{{\bm{L}}}
\def\mM{{\bm{M}}}

\def\mP{{\bm{P}}}

\def\mR{{\bm{R}}}

\def\mU{{\bm{U}}}

\def\mW{{\bm{W}}}

\def\mY{{\bm{Y}}}

\DeclareMathAlphabet{\mathsfit}{\encodingdefault}{\sfdefault}{m}{sl}
\SetMathAlphabet{\mathsfit}{bold}{\encodingdefault}{\sfdefault}{bx}{n}

\def\sD{{\mathbb{D}}}

\def\sI{{\mathbb{I}}}

\def\sO{{\mathbb{O}}}

\def\sR{{\mathbb{R}}}

\DeclareMathOperator{\E}{\mathbb{E}}  

\DeclareMathOperator{\fvec}{vec}

\definecolor{cvprblue}{rgb}{0.21,0.49,0.74}
\usepackage[pagebackref,breaklinks,colorlinks,allcolors=cvprblue]{hyperref}

\def\paperID{4} 
\def\confName{CVPR}
\def\confYear{2026}

\title{Certified Training for Convolutional Perturbations}

\author{Benedikt Brückner\\
Safe Intelligence\\
{\tt\small benedikt@safeintelligence.ai}
\and
Alessio Lomuscio\\
Safe Intelligence\\
Imperial College London\\
{\tt\small alessio@safeintelligence.ai}
}

\begin{document}
\maketitle
\begin{abstract}
Vision models have been found to be susceptible to
perturbations such as motion blur induced at runtime by a shaking camera.
This impedes their deployment in critical applications since phenomena such
as slightly blurred vision might lead to failures, for example an object
detector missing objects. While methods such as data augmentation or
Adversarial Training can improve empirical robustness, they lack formal
safety guarantees, making it difficult to identify and mitigate hidden
vulnerabilities. We introduce a novel Certified Training approach that
leverages an efficient encoding of convolutional perturbations to train
provably robust models. Our method significantly outperforms Adversarial
Training, achieving, for example, over $80\%$ robust accuracy against
motion blur of reasonable intensity on CIFAR10 while maintaining comparable
standard accuracy.
\end{abstract} 
\section{Introduction} \label{sec:introduction}
Neural networks trained for image processing tasks are often brittle and
may produce unexpected results under slight changes of their inputs
\cite{Szegedy+14}, including perturbations such as motion blur which
frequently appear in real-world systems \cite{Guo+20}. This impedes the
deployment of such models in safety-critical domains such as self-driving
cars and aviation. While additional training data and methods such as
Adversarial Training can improve the generalisation capabilities of the
model being trained, their safety usually cannot be verified. Since even
models which are thought to be robust may have unforeseen weaknesses and
blind spots \cite{CarliniWagner17a} , methods for verifying the robustness
to input perturbations are needed. To this end, a number of approaches for
verifying neural networks
\cite{Katz+17,Liu+20a,Ferrari+22,Zhang+22,Zhou+24,Kouvaros+25} have been put forward.
Many works focus on perturbations in the $l_\infty$ norm, often referred to
as white noise, others consider perturbations such as brightness changes
\cite{KouvarosLomuscio18}, rotations \cite{Singh+19a} or
occlusions~\cite{Guo+23}. However, models trained using standard training
approaches are usually difficult to verify. This is due to the looseness of
the convex relaxations which verifiers employ to model nonlinear
dependencies \cite{Mao+24}. Certified Training methods aim at training
verifiably robust networks, often by adding a loss component based on
bounds on the model's outputs to the training
loss~\cite{Li+20a,Muller+22,DePalma+24,Mao+24}. Despite the fact that
convolutional perturbations such as motion blur frequently occur in
safety-critical systems like autonomous aircraft or self-driving cars~
\cite{OktayCelikTurkmen18,SayedBrostow21,Guo+20}, no method for training
neural networks that are verifiably robust to such convolutional
perturbations exist. We fill this gap through the following contributions:
\begin{itemize}
\item We present the first method for training models that are certifiably
robust to motion blur and related convolutional perturbations. Exploiting
the low perturbation dimensionality together with an efficient, batched
implementation, we compute tight bounds for training. This significantly
reduces the overregularisation of the model to arrive at a sweet spot
between standard accuracy and robustness.
\item We conduct a comprehensive evaluation across perturbations and
corruption strengths. For reasonably sized perturbations we achieve much
higher robust accuracies than pure Adversarial Training while standard
performance remains comparable. Even for strong perturbations we obtain
models with a robust accuracy of over $70\%$ in the majority of cases
for CIFAR10. Standard and robust accuracies are often well above those of
the state-of-the-art Certified Training methods for white noise.
\end{itemize}

\section{Related Work}
\label{sec:related_work}
\paragraph{Neural Network Verification} Neural networks are well known to be
susceptible to adversarial attacks
\cite{Szegedy+14,Papernot+16,Eykholt+18,Liu+19b,Tu+20}. Algorithms for
verifying neural networks prove the nonexistence of adversarial examples
within the neighbourhood of a given input using standard solvers
\cite{Katz+17,Singh+18} or custom bound propagation procedures~
\cite{Liu+20a,Wang+18a,Wang+18b,Singh+19a,Xu+21,Wang+21b}. While such methods
predominantly target image classifiers, robustness verification has also been
extended to other tasks such as object detection \cite{Cohen+24,Brueckner+26}.
Of special interest to this work are verifiers based on propagating
intervals through a model to bound the output neurons. Interval Bound
Propagation (IBP) cheaply propagates concrete bounds through the network. Alternatives such as
(Forward) Symbolic Interval Propagation (SIP) or
CROWN/back-substitution/RSIP exploit symbolic bounding equations to gain
precision \cite{Wang+18a,Wang+18b,Singh+19a,Xu+21,Wang+21b}.
\paragraph{Perturbation Types}
Most works investigate robustness to white noise, represented by pixel-wise
perturbations bounded in the $l_\infty$ norm~\cite{Singh+18,Katz+19}.
Further perturbations of interest include brightness and contrast changes
\cite{KouvarosLomuscio18,Henriksen+21}, geometric perturbations such as
translations or rotations~\cite{Singh+19a,Balunovic+19a,Batten+24} as well
as occlusions \cite{Mohapatra+20}. More complex perturbations may be
modeled using generative models
\cite{Mirman+21,HanspalLomuscio23,Waite+23}. Prior work on convolutional
perturbations, which this work focuses on, is scarce. \citet{Guo+20} show that
motion blur in input images degrades the performance of neural networks,
but there is little work on evaluating and verifying the robustness of
models to such phenomena
\cite{Paterson+21,MziouSallamiAdjed22,BruecknerLomuscio25}.
\citet{Paterson+21} perform testing for blur perturbations while
\citet{MziouSallamiAdjed22} develop a method for verification against
general convolutional perturbations. Since this method faces scalability
issues, \citet{BruecknerLomuscio25} suggest to instead focus on verifying
robustness to specific convolutional perturbations which enables scaling to
larger models. However, both works on convolutional perturbations focus on
robustness verification and not training. This raises the question of how
the verified robustness of brittle networks can be improved.
\paragraph{Adversarial and Certified Training}
A problem closely related to robustness verification is that of training
networks to be more robust. Adversarial Training increases the empirical
robustness \cite{Madry+18,GoodfellowShlensSzegedy15} while Robust/Verified/Certified
Training improves the verified robustness of a model. This can be
achieved by adding regularisation to Adversarial Training
\cite{Xiao+19a,Palma+22} or computing an upper bound on the worst case loss
\cite{Gowal+19,WongKolter17,Shi+21a}. State-of-the-art (SoA) approaches for
robustification to white noise combine over- and underapproximations of the
true worst-case loss \cite{Muller+22,Mao+23,DePalma+24}, making use of IBP
for bounding.
\citet{HenriksenLomuscio23} focus on certified training for low-dimensional
bias field perturbations. Since the convolutional perturbation encoding we
focus on is also low-dimensional \cite{BruecknerLomuscio25}, we draw on
their insights and also use symbolic bounds to compute upper bounds of the
loss.

\section{Background}
\label{sec:background}
We use bold lower-case letters $\va$ to denote vectors and bold upper-case
letters $\mA$ to denote matrices and tensors. Square brackets are used for
indexing vectors, matrices and tensors with \eg $\mA[i, j]$ denoting the
element in row $i$ and column $j$ of the matrix $\mA$ and $\mA[i, :]$ the
$i$-th row of $\mA$. We write $\mI*\mK$ for the convolution of a matrix
$\mI$ with a kernel $\mK$. $\fvec(\mA)$ is used for the vectorisation of a
matrix or tensor, \eg if $\mA \in \sR^{m \times n}$ and $\va = \fvec(\mA)$
then $\va \in \sR^{mn}$. The inverse operation is $\fvec^{-1}$, to clarify
the shape of the output may be added. For the above example one would write
$\fvec^{-1}_{m, n}(\va)$ to restore the original $\mA$.
We focus on feed-forward neural networks $f: \mathbb{R}^{n_0} \rightarrow
\mathbb{R}^{n_L}$ with $L$ layers which operate on an input $\vx \in \mathbb{R}^{n_0}$.
We restrict our analysis to networks with ReLU activations, but note
that both verification and certified training are equally feasible for
networks incorporating other activation functions \cite{Wu+23,Shi+24}.

\subsection{Neural Network Verification}
\label{ssec:neural_network_verification} Neural Network Verification aims
at verifying the robustness of a trained neural network $f$ to
perturbations of its input. For a set of inputs $\sI \subset
\mathbb{R}^{n_0}$ and a safe output set $\sO \subset \mathbb{R}^{n_L}$,
the aim is to either prove or disprove that $\forall \vx \in \sI: f(\vx)
\in \sO$. The verification literature usually considers perturbations of a
given strength $\epsilon$ bounded in the $l_\infty$ norm.
To determine whether a network is robust, most SoA verifiers make use of
bound propagation methods.\\
Interval Bound Propagation
\cite{Gowal+19,Wang+18b} is the cheapest and least precise such method
which propagates concrete bounds through the network.
Let $\mW_i, \vb_i$ denote the weight matrix and bias vector for the $i$-th
layer and $\mW_i^+, \mW_i^-$ the matrix containing only the
positive/negative elements of $\mW_i$ and $0$ everywhere else. Given
concrete lower and upper bounds $\vl_{i-1}, \vu_{i-1}$ for a layer, the
bounds for the following layer are calculated as $\vl_i = \mW_i^+ \vl_{i-1}
+ \mW_i^- \vu_{i-1} + \vb_i$ and $\vu_i = \mW_i^+ \vu_{i-1} + \mW_i^-
\vl_{i-1} + \vb_i$.
Symbolic Bound Propagation methods alleviate the coarseness of IBP by
propagating symbolic equations through the network. This tracks dependencies
between neurons, leading to notably tighter bounds.
Forward SIP (FSIP)/Standard SIP (SSIP) performs these computations in a
forward manner, from the input to the output layer
\cite{Wang+18a,KernBueningSinz22}. Even tighter bounds can be obtained by
bounding from the layer of interest in a reversed manner.
This method is referred to as either back-substitution, CROWN
\cite{Zhang+18}, DeepPoly \cite{Singh+19a} or Reversed Symbolic Interval
Propagation (RSIP) \cite{HenriksenLomuscio21}, and can be performed
dynamically to improve efficiency \cite{Kouvaros+25}.
Combining these bounding algorithms with a branching strategy yields
subproblems with tighter approximations
\cite{Botoeva+20,DePalma+21,LanBruecknerLomuscio23}.

\subsection{Adversarial and Certified Training}
\label{ssec:adversarial_certified_training}
A neural network $f_\theta: \sR^{n_0} \rightarrow \sR^{n_L}$ with
parameters $\theta$ is trained on a dataset $\sD$ containing pairs of
inputs and labels $(\vx, \vy)$ by solving the following optimisation
problem
\begin{equation}
\min_\theta \displaystyle \mathop{\E}_{(\vx, \vy) \sim \sD} \loss \left (
f_\theta \left ( \vx \right ), \vy \right )
\end{equation}
Networks trained by using this standard training loss are easily attackable
using various adversarial attack paradigms since the above formulation does
not induce stability of the model around a given input in any way. Let
$\sI(\vx, \epsilon)$ denote a set of admissible perturbations of a clean
input $\vx$. \citet{Madry+18} define $\sI(\vx, \epsilon) = \left \{
\tilde{\vx} \in \mathbb{R}^{n_0} \mid \| \tilde{\vx} - \vx \|_\infty \leq
\epsilon \right \}$ and suggest to instead solve
\begin{equation}
\min_\theta \displaystyle \mathop{\E}_{(\vx, \vy) \sim \sD} \: \max_{\tilde{x}
\in \sI(\vx, \epsilon)} \loss \left ( f_\theta \left ( \tilde{\vx} \right ), \vy
\right )
\end{equation}
Maximising $\loss \left ( f_\theta \left ( \vx \right ), \vy \right )$ is a
nonconvex optimisation problem. Instead of solving it exactly, Adversarial
Training (AT) finds samples that locally maximise the inner problem by using \eg
Projected Gradient Descent (PGD) \cite{Madry+18}. Computing the loss on
these yields a lower bound for the true maximum
which is used to update the model. This makes it more difficult to
find adversarial examples using attacks (increases the empirical
robustness), but often leads to decreased standard accuracies and the
trained model is usually still difficult to verify. Also, adversarial examples
can usually still be found, for example using stronger attacks.
Certified Training computes a sound upper bound for the inner
problem using bounding algorithms such as IBP or SSIP and then performs
gradient descent on this upper bound. For details we refer to
\cite{Gowal+19}. Certified training induces a strong regularisation effect,
achieving increased stability of the trained model and
facilitating verification at the expense of a decreased standard accuracy.
Most importantly, it usually produces models for which we can \textit{prove}
that there exist \textit{no incorrectly classified inputs in the neighbourhood of a
given input.}

Early methods use IBP bounds or combinations of IBP and tighter
CROWN bounds for training \cite{MirmanGehrVechev18a,Zhang+19} but require
long training schedules and warm-up phases during which the perturbation
size is ramped up to achieve satisfying performance. \citet{Shi+21a} reduce
the length of the training schedules by adding batch normalisation (BN) layers
to the network \cite{IoffeSzegedy15}, developing a custom weight
initialisation called IBP Initialisation and adding a warmup loss during
the warmup phase to stabilise bound tightness and activation states. Recent
work on Certified Training for white noise uses IBP bounds as a sound
overapproximation, but combines them with an Adversarial Training component
\cite{Muller+22,Mao+23,DePalma+24}.
For low-dimensional perturbations such as brightness perturbations, the
changes of pixels' values are coupled across an image, leading to tighter
bounds compared to a white noise perturbation where each pixel can vary
independently. \citet{HenriksenLomuscio23} show that Certified Training
using IBP is ineffective for low dimensional perturbations due to the loss
of dependency information in IBP. This can be mitigated by instead using
SSIP or RSIP for bounding which produce significantly tighter bounds. One
drawback of RSIP is the fact that a separate bound computation pass needs
to be initiated for every activation layer. \citet{HenriksenLomuscio23}
introduce RSIP-SSIP as a cheaper alternative which first
uses SSIP to calculate the bounds for all layers in the network and then
uses one final RSIP pass to obtain tight bounds on the robust loss.\\

\subsection{Convolutional Perturbations}
\label{ssec:convolutional_perturbations}
The discrete convolution of an image $\mI$ with a kernel $\mK$ enables a
variety of effects such as blurring or sharpening to be applied to an input
\cite[pp.~95--99]{Szeliski22}. Such convolutional perturbations of an image are likely to
appear in the real world, testing the performance of a system on
perturbed images or attacking it using noise models such as adversarial
blurring \cite{Guo+20} can give an indication of its robustness. Still,
neither of these techniques can provide guarantees that a model is actually
robust. \citet{MziouSallamiAdjed22} develop a method for verifying the
robustness of a model to convolutional perturbations
but it provides loose bounds and struggles to scale to larger networks.
Parameterised kernels instead model convolutional
perturbations in a way that enables scaling to larger networks while still
proving robustness for dense input regions containing an infinite number of
instantiations of the perturbation \cite{BruecknerLomuscio25}. They use a
parameter $z \in [0, 1]$ to linearly interpolate between an identity kernel
and a perturbation kernel. In the simple example of box blurring with a
kernel size of $s=3$, the parameterised kernel must satisfy
\begin{equation}
\mP_{z=0} = \begin{pmatrix}
0 & 0 & 0 \\[2pt]
0 & 1 & 0 \\[2pt]
0 & 0 & 0 \\
\end{pmatrix},
\mP_{z=1} = \begin{pmatrix}
\frac{1}{9} & \frac{1}{9} & \frac{1}{9} \\[2pt]
\frac{1}{9} & \frac{1}{9} & \frac{1}{9} \\[2pt]
\frac{1}{9} & \frac{1}{9} & \frac{1}{9} \\
\end{pmatrix}.
\end{equation}
These equations can be used to derive a linear parameterisation $\mP_z =
\mC z + \mD$ where $\mC, \mD, \mP \in \sR^{3 \times 3}$ and $z \in \sR$.
Due to the fact that the identity kernel is not well-defined for even
kernel sizes \cite{BruecknerLomuscio25}, we only consider perturbation
kernels with odd sizes $s$.
Let $\mI$ be an image with $\mI \in
\sR^{o_c \times o_h \times o_w}$ where $o_c$ the number of channels of the
image, $o_h$ the image height and $o_w$ the image width. Increasing $z$
from $0$ to $1$ increases the strength of the perturbation applied to $\mI$
when it is convolved with it. For an image $\mI$ and an arbitrary
parameterised kernel $\mK = \mA z + \mB$ with $\mK, \mA, \mB \in \sR^{s
\times s}$ where $s$ is the size of the perturbation kernel, it can be
shown that \cite[Theorem 1]{BruecknerLomuscio25}
\begin{equation}
\mI * \mK = \underbrace{\left ( \mI * \mA \right )}_{\mR_\mA} z + \underbrace{\left ( \mI * \mB \right )}_{\mR_\mB} \label{eq:separable_conv}
\end{equation}
The result of the two convolutions $\mR_\mA, \mR_\mB$ can therefore be
computed separately. The result of the perturbation of $\mI$ with $\mK$ for
an arbitrary $z \in [0, 1]$ can then be obtained by simply computing
$\mI_\text{perturbed} = \mR_\mA \cdot z + \mR_\mB$. No new convolution
needs to be run for this, only the multiplication of $\mR_\mA$ with the
scalar $z$ is required before adding the result to $\mR_\mB$. This trick
can be used to encode convolutional perturbations for arbitrary kernels
$\mK$ into a verification framework which may then be used to certify
robustness for any perturbation strength within a dense interval $[0,
\epsilon]$.

\section{Method}
\label{sec:method}
\def\adjboxvcenter{0.2\baselineskip}
\begin{figure}[!tb]
\centering
\begin{center}
\SetTblrInner{rowsep=2pt,colsep=2pt}
\begin{tblr}{
        measure=vbox,
        width=\linewidth,
        colspec={X[l,c,m] X[2,c,m] X[2,c,m] X[2,c,m] X[2,c,m]},
    }
    {} & {\small $\eps=0.0$} & {\small $\eps=0.2$} & {\small $\eps=0.6$} & {\small $\eps=1.0$} \\
    {
        \small Box Blur
    } &
    \adjustbox{valign=m}{\includegraphics[width=4em, height=4em]{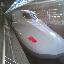}}
    &
    \adjustbox{valign=m}{\includegraphics[width=4em, height=4em]{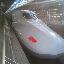}}
    &
    \adjustbox{valign=m}{\includegraphics[width=4em, height=4em]{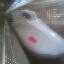}}
    &
    \adjustbox{valign=m}{\includegraphics[width=4em, height=4em]{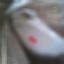}}\\
    {
        \small Sharpen
    } &
    \adjustbox{valign=m}{\includegraphics[width=4em, height=4em]{figures/perturbation_strength_visualisation/before_perturbation.jpg}}
    &
    \adjustbox{valign=m}{\includegraphics[width=4em, height=4em]{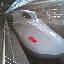}}
    &
    \adjustbox{valign=m}{\includegraphics[width=4em, height=4em]{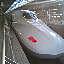}}
    &
    \adjustbox{valign=m}{\includegraphics[width=4em, height=4em]{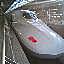}}\\

    {
        \small Motion Blur,\\
        $0^\circ$
    } &
    \adjustbox{valign=m}{\includegraphics[width=4em, height=4em]{figures/perturbation_strength_visualisation/before_perturbation.jpg}}
    &
    \adjustbox{valign=m}{\includegraphics[width=4em, height=4em]{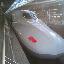}}
    &
    \adjustbox{valign=m}{\includegraphics[width=4em, height=4em]{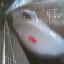}}
    &
    \adjustbox{valign=m}{\includegraphics[width=4em, height=4em]{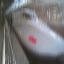}}
\end{tblr}
\end{center}
\caption{Visualisation of different perturbations and their strengths $\epsilon$ on a $64 \times 64$ image from the ``Bullet Train'' class in TinyImageNet using a kernel size of $s=5$.}
\label{fig:ablation_study_ibp_init_warmup}
\end{figure}
Training verifiably robust models with high standard accuracies is a
challenging problem since Standard and Adversarial Training as well as
training using approaches such as data augmentation produce models that are
still vulnerable to corruptions of the inputs. Certified Training is often
slow, may suffer from loose bounds and risks overregularising the model.
Besides this, IBP-based Certified Training is ineffective for robustifying
models against low-dimensional perturbations \cite{HenriksenLomuscio23}. We
design a Certified Training scheme for convolutional perturbations that
makes use of SSIP and RSIP bounds to effectively keep track of the
low-dimensional structure of the perturbations at hand. By exploiting this
low dimensionality instead of relying on high-dimensional encodings
producing looser bounds \cite{MziouSallamiAdjed22}, we avoid
overregularisation of the models and preserve a high standard accuracy on
the tasks at hand. Together with our efficient, batched perturbation
encoding we obtain a highly effective framework for training models that
are verifiably robust to convolutional perturbations.
To encode the desired perturbations, we prepend suitable layers to the
model being trained \cite{KouvarosLomuscio18,Mohapatra+20}. Certified
Training poses additional challenges to the prepended encodings since they
need to support batched processing of the inputs. We assume that for a
training step we are given a batch of images $\mI \in \sR^{o_b \times o_c
\times o_h \times o_w}$ where $o_b$ is the batch size. We use the
parameterised perturbation kernels described by \citet{BruecknerLomuscio25}
to simulate perturbations such as motion blur or sharpen. These can be
written as $\mK = \mA z + \mB$ with $z \in \sR$ as well as $\mK, \mA, \mB
\in \sR^{s \times s}$. To obtain the linear coefficients for encoding the
perturbation in a parameterised way, we leverage equation
\ref{eq:separable_conv}. Using a batched convolution operation, the
coefficient and bias tensor $\mR_\mA, \mR_\mB$ for the entire batch are
computed by convolving each input in the batch with $\mA, \mB$. Reflect
padding is used to ensure that $\mR_\mA, \mR_\mB \in \sR^{o_b \times o_c
\times o_w \times o_h}$.

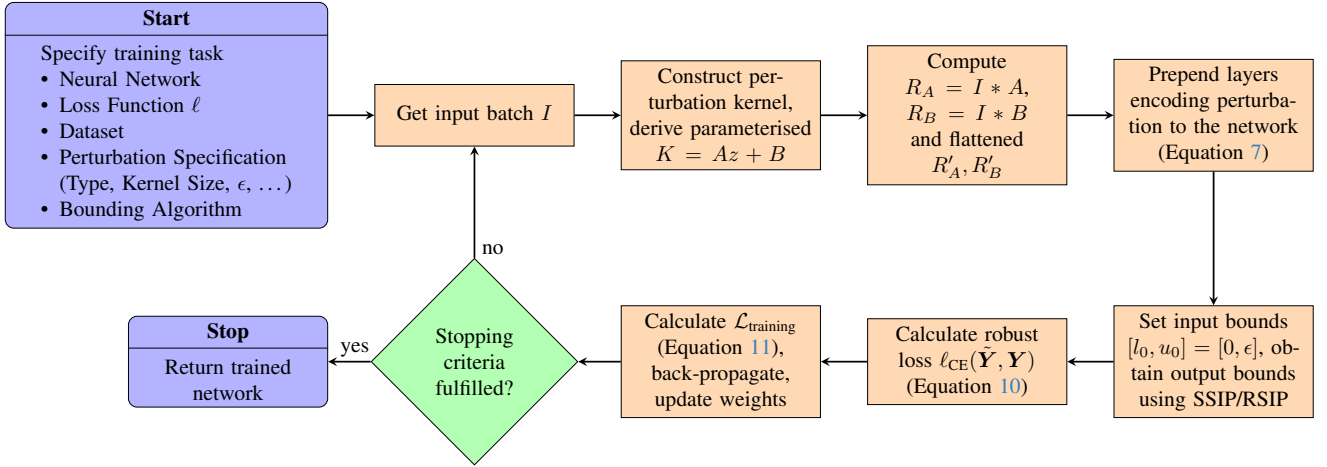
\begin{figure*}[!htb]
\centering
\usetikzlibrary{shapes.geometric, arrows, backgrounds}
\tikzset{
  font={\fontsize{10pt}{12}\selectfont}
}
\tikzstyle{io} = [rectangle, rounded corners, 
minimum width=3cm, 
minimum height=1cm,
text centered,
text width=5cm,
draw=black, 
fill=blue!30]

\tikzstyle{process} = [rectangle, 
minimum width=3cm, 
minimum height=1cm, 
text centered, 
text width=3cm, 
draw=black, 
fill=orange!30]

\tikzstyle{decision} = [diamond, 
minimum width=3cm, 
minimum height=1cm, 
text centered,
text width=1.8cm,
draw=black, 
fill=green!30]
\tikzstyle{arrow} = [thick,->,>=stealth]
\resizebox{\linewidth}{!}{
\begin{tikzpicture}[node distance=2cm]

\node (in) [io] {
\textbf{Start} \\[3mm]
\begin{varwidth}{\linewidth}
Specify training task
\begin{itemize}
        \item Neural Network
        \item Loss Function $\ell$
        \item Dataset
        \item Perturbation Specification \\(Type, Kernel Size, $\epsilon$, \dots)
        \item Bounding Algorithm
    \end{itemize}\end{varwidth}
};
\node (proc1) [process, right of=in, xshift=3cm] {Get input batch $I$};

\node (proc2) [process, right of=proc1, xshift=2cm] {Construct perturbation kernel, derive parameterised $K = A z + B$};

\node (proc3) [process, right of=proc2, xshift=2cm] {Compute\\ $R_A = I * A$,\\ $R_B = I * B$\\ and flattened $R_A', R_B'$};

\node (proc4) [process, right of=proc3, xshift=2cm] {Prepend layers encoding perturbation to the network (Equation \ref{eq:batched_operation})};

\node (proc5) [process, below of=proc4, yshift=-2cm] {Set input bounds $[l_0, u_0] = [0, \epsilon]$, obtain output bounds using SSIP/RSIP};

\node (proc6) [process, left of=proc5, xshift=-2cm] {Calculate robust loss $\ell_\text{CE}(\tilde{\mY}, \mY)$ (Equation \ref{eq:robust_loss})};

\node (proc7) [process, left of=proc6, xshift=-2cm] {Calculate $\mathcal{L}_\text{training}$ (Equation \ref{eq:total_loss}), back-propagate, update weights};

\node (dec) [decision, left of=proc7, xshift=-2cm] {Stopping criteria \\ fulfilled?};

\node (out) [io, left of=dec, xshift=-2cm, text width=3cm] {\textbf{Stop} \\[1.5mm] Return trained network};

\draw [arrow] (in) -- (proc1);
\draw [arrow] (proc1) -- (proc2);
\draw [arrow] (proc2) -- (proc3);
\draw [arrow] (proc3) -- (proc4);
\draw [arrow] (proc4) -- (proc5);
\draw [arrow] (proc5) -- (proc6);
\draw [arrow] (proc6) -- (proc7);
\draw [arrow] (proc7) -- (dec);

\draw [arrow] (dec) -- node[anchor=north,pos=.2,xshift=3mm] {no} (proc1);
\draw [arrow] (dec) -- node[anchor=west,pos=.9,yshift=2mm] {yes} (out);

\draw ([yshift=13mm]in.west) -- ([yshift=13mm]in.east);
\draw ([yshift=2.2mm]out.west) -- ([yshift=2.2mm]out.east);

\end{tikzpicture}
}
\caption{A flowchart showing the certified training pipeline that we propose. Stopping criteria can, for example, be a maximum number of epochs or a (robust) accuracy threshold being surpassed.}
\label{fig:flowchart}
\end{figure*}

Let $d = o_c \cdot o_h \cdot o_w$ be
the number of pixels in each image in a batch. After running the
pre-computation for a batch, the obtained tensors are transformed to
\begin{align}
\mR_\mA' &= \fvec^{-1}_{o_b, d, 1} \left ( \fvec \left ( \mR_\mA \right ) \right ) \label{eq:tensor_transform_1} \\
\mR_\mB' &= \fvec^{-1}_{o_b, d} \left ( \fvec \left ( \mR_\mB \right ) \right ) \label{eq:tensor_transform_2}
\end{align}
This is equivalent to flattening each convolution result in a batch to
become a vector with a trailing dimension of size one. A reshape layer that
encodes the linear operation based on $\mR_\mA', \mR_\mB'$ is prepended to
the network. This layer takes the input to the model $\tilde{\vz} \in
\sR^{o_b}$ and applies the linear mapping to each entry in $\tilde{\vz}$.
Here, the $i$-th entry $\tilde{\vz}[i]$ controls the degree to which the
$i$-th image in the batch is perturbed. Using Einstein notation, the
batched operation can be written as
\begin{equation}
\mI_\text{perturbed}'[i, j] = \sum_k \left ( \mR_\mA'[i, j, k] \tilde{z}[i, k] \right ) + \mR_\mB'[i, j] \label{eq:batched_operation}
\end{equation}
To ultimately obtain the batch of perturbed images,
\begin{equation}
\mI_\text{perturbed} = \fvec^{-1}_{o_b, o_c, o_h, o_w} \left ( \mI_\text{perturbed}' \right ) \label{eq:tensor_transform_3}
\end{equation}
is computed. This tensor then serves as the input for the actual neural
network to be trained.
The convolution of the input can be computed in a parallelised manner as a
preprocessing step and is run once for every batch. All other operations
(partial flattening, batched multiplication, reconstruction of the
correctly shaped perturbed images) are linear, thus differentiable. When
these layers are prepended to a network, they support forward passes, bound
propagation and backpropagation of gradients through them. For SSIP and
RSIP-SSIP based training we use the following approach after prepending the
layers encoding the perturbation according to Equations
\ref{eq:batched_operation} and \ref{eq:tensor_transform_3}: The bounds for
the first layer are set to $[\mL_0, \mU_0] = [0, \epsilon]^{o_b}$. Using
the chosen bounding algorithm, we obtain bounds $[\mL_L, \mU_L]$ for the
final ($L$-th) layer. Let $\mY \in \{0, 1\}^{o_b \times c}$ be the one-hot
encoding of the labels $y$ of a given batch of images $\mI$. Since $y[i]$
is the correct class of the $i$-th input, we then define
\begin{equation} \label{eq:worst_case_ce_loss}
\tilde{\mY}[i, j]=
\begin{cases}
\mL_L[i, j], \text{if}\: j = y[i]\\
\mU_L[i, j], \text{otherwise}
\end{cases}
\end{equation}
and can use this new matrix in the cross-entropy loss function
$\ell_\text{CE}$ to calculate the robust loss as
\begin{equation}
\mathcal{L}_\text{method} = \ell_\text{CE}(\tilde{\mY}, \mY) \label{eq:robust_loss}
\end{equation}
where $\text{method} \in \{\text{SSIP}, \text{RSIP-SSIP}\}$
\cite{Gowal+19}. For Adversarial Training we compute the Projected Gradient
Descent (PGD) loss $\mathcal{L}_\text{PGD}$ as follows: We randomly
initialise the attack point in the $[0, \epsilon]^{o_b}$ interval,
calculate the loss and its gradients with respect to the input point,
update the attack point by taking a step in the direction of the sign of
the gradient and repeat this $8$ times.
We use Adversarial Training as a baseline
to compare the standard accuracy of our method to that of a method
inducing a weaker regularisation effect. However, just like approaches such
as data augmentation or Adversarial Training based on other attack schemes,
PGD training does not improve the certified accuracy of a network. As
mentioned in Section \ref{ssec:adversarial_certified_training}, the current
SoA certified training methods are based on loose IBP bounds which
overregularise networks when used in the loss function, leading to
decreased standard accuracies. The SoA methods introduce an
underapproximation component into their loss formulation to prevent this.
By using tighter bounds from the SSIP and the RSIP-SSIP method, we
circumvent these issues without relying on similar tricks. Our framework
retains good standard accuracies while achieving much increased verified
accuracies. Denoting the value of the standard cross-entropy loss as
$\mathcal{L}_\text{CE}$ and the robustification loss as
$\mathcal{L}_\text{method}$ where $\text{method} \in \{\text{PGD},
\text{SSIP}, \text{RSIP-SSIP}\}$, our loss function becomes

\begin{equation}
\mathcal{L}_\text{training} = \frac{\mathcal{L}_\text{CE} + \mathcal{L}_\text{method}}{2} \label{eq:total_loss}
\end{equation}
and we can use a standard machine learning framework to train networks using this loss.
\section{Evaluation}
\label{sec:evaluation}
\begin{table*}[!htb]
\small
\centering
\adjustbox{max width=\textwidth}{%
\begin{tabular}{cccccccccccccccc} 
\toprule  
\multicolumn{4}{c}{} & \multicolumn{4}{c}{Standard Acc.} & \multicolumn{4}{c}{Empirical Acc.} & \multicolumn{4}{c}{Robust Acc.} \\
\cmidrule(r){5-8} \cmidrule(r){9-12} \cmidrule(r){13-16}
Dataset & Perturbation & $s$ & $\epsilon$ & STD & PGD & SSIP & RS & STD & PGD & SSIP & RS & STD & PGD & SSIP & RS \\
\cmidrule(lr){1-4} \cmidrule(lr){5-16}
 & & \multirow{3}{*}{3} & 0.2 & \textbf{91.35} & 88.23 & 89.45 & 90.49 & \textbf{90.05} & 86.88 & 88.54 & 89.82 & 0.00 &  0.00 & \textbf{87.43} & \textbf{87.43} \\
 & & & 0.6 & \textbf{91.35} & 87.97 & 85.14 & 87.31 & 79.09 & 84.33 & 83.15 & \textbf{85.26} & 0.00 & 0.00 & 79.90 & \textbf{80.02}  \\
 & & & 1.0 & \textbf{91.35} & 87.08 & 81.13 & 83.10 & 46.80 & 80.63 & 78.70 & \textbf{80.76} & 0.00 & 0.00 & 74.25 & \textbf{74.33}  \\ 
\cmidrule(r){3-16}
 & & \multirow{3}{*}{5} & 0.2 & \textbf{91.35} & 88.12 & 88.99 & 90.60 & 89.81 & 86.47 & 88.27 & \textbf{89.93} & 0.00 & 0.00 & 86.88 & \textbf{87.01} \\
 & Box Blur & & 0.6 & \textbf{91.35} & 87.91 & 84.79 & 87.27 & 73.04 & 84.21 & 81.78 & \textbf{84.22} & 0.00 & 0.00 & 74.76 & \textbf{76.10} \\
 & & & 1.0 & \textbf{91.35} & 86.26 & 73.52 & 76.42 & 37.46 & \textbf{74.40} & 68.68 & 71.46 & 0.00 & 0.00 & 62.46 & \textbf{62.58} \\ 
\cmidrule(r){3-16}
 & & \multirow{3}{*}{7} & 0.2 & \textbf{91.35} & 88.22 & 89.14 & 90.3 & \textbf{89.65} & 86.76 & 88.34 & 89.62 & 0.00 & 0.00 & \textbf{87.00} & 86.74 \\
 & & & 0.6 & \textbf{91.35} & 87.64 & 84.98 & 87.31 & 74.87 & 83.47 & 81.74 & \textbf{84.79} & 0.00 & 0.00 & 74.46 & \textbf{76.88} \\
 & & & 1.0 & \textbf{91.35} & 84.64 & 69.91 & 72.73 & 32.72 & \textbf{69.10} & 61.64 & 64.93 & 0.00 & 0.00 & 54.12 & \textbf{56.31} \\ 
\cmidrule(r){2-16}
 & & \multirow{3}{*}{3} & 0.2 & \textbf{91.35} & 88.42 & 89.08 & 90.40 & \textbf{90.75} & 87.69 & 88.48 & 89.89 & 0.00 & 0.02 & 87.88 & \textbf{87.99} \\
 & & & 0.6 & \textbf{91.35} & 88.56 & 87.68 & 89.48 & 87.48 & 87.07 & 86.61 & \textbf{88.51} & 0.00 & 0.00 & \textbf{84.63} & 84.39 \\
 & & & 1.0 & \textbf{91.35} & 88.50 & 86.92 & 88.05 & 84.35 & 86.16 & 85.83 & \textbf{86.93} & 0.00 & 0.00 & \textbf{82.54} & 81.11 \\ 
\cmidrule(r){3-16}
 & & \multirow{3}{*}{5} & 0.2 & \textbf{91.35} & 88.82 & 89.32 & 90.41 & \textbf{90.54} & 87.90 & 88.64 & 89.72 & 0.00 & 0.00 & \textbf{87.52} & 87.43 \\
CIFAR10 & Sharpen & & 0.6 & \textbf{91.35} & 88.75 & 88.42 & 90.06 & 87.01 & 86.75 & 87.01 & \textbf{88.89} & 0.00 & 0.00 & 83.62 & \textbf{84.22} \\
 & & & 1.0 & \textbf{91.35} & 88.30 & 87.52 & 89.38 & 82.34 & 85.80 & 85.59 & \textbf{87.82} & 0.00 & 0.00 & 80.94 & \textbf{81.91} \\ 
\cmidrule(r){3-16}
 & & \multirow{3}{*}{7} & 0.2 & \textbf{91.35} & 88.71 & 89.39 & 90.62 & \textbf{90.16} & 87.75 & 88.75 & 90.01 & 0.00 & 0.00 & 87.58 & \textbf{87.60} \\
 & & & 0.6 & \textbf{91.35} & 88.81 & 88.10 & 89.97 & 87.00 & 87.00 & 86.60  & \textbf{88.71} & 0.00 & 0.00 & 83.58 & \textbf{84.21} \\
 & & & 1.0 & \textbf{91.35} & 88.01 & 87.07 & 89.00 & 82.18 & 85.19 & 84.89 & \textbf{87.23} & 0.00 & 0.00 & 80.99 & \textbf{82.38} \\ 
\cmidrule(r){2-16}
 & & \multirow{3}{*}{3} & 0.2 & \textbf{91.35} & 88.19 & 89.00 & 89.87 & \textbf{90.24} & 87.23 & 88.49 & 89.39 & 0.00 & 0.06 & \textbf{88.05} & 87.92 \\
 & & & 0.6 & \textbf{91.35} & 88.01 & 86.94 & 88.20 & 85.10 & 85.98 & 86.00 & \textbf{87.18} & 0.00 & 0.00 & \textbf{84.90}  & 84.27 \\
 & & & 1.0 & \textbf{91.35} & 87.77 & 85.69 & 86.94 & 75.68 & 84.28 & 84.40 & \textbf{85.80} & 0.00 & 0.00 & \textbf{82.84} & 82.02 \\ 
\cmidrule(r){3-16}
 & & \multirow{3}{*}{5} & 0.2 & \textbf{91.35} & 88.06 & 89.13 & 90.19 & \textbf{90.09} & 87.02 & 88.22 & 89.51 & 0.00 & 0.00 & 86.91 & \textbf{87.00} \\
 & Motion Blur & & 0.6 & \textbf{91.35} & 88.19 & 84.80  & 85.92 & 76.93 & \textbf{84.87} & 83.01 & 84.08 & 0.00 & 0.00 & \textbf{80.53} & 79.39 \\
 & & & 1.0 & \textbf{91.35} & 86.51 & 82.03 & 83.93 & 51.78 & 80.79 & 80.22 & \textbf{81.92} & 0.00 & 0.00 & \textbf{76.58} & 75.96 \\ 
\cmidrule(r){3-16}
 & & \multirow{3}{*}{7} & 0.2 & \textbf{91.35} & 88.49 & 89.06 & 90.36 & \textbf{89.86} & 87.05 & 87.96 & 89.66 & 0.00 & 0.00 & 86.38 & \textbf{86.55} \\
 & & & 0.6 & \textbf{91.35} & 88.65 & 84.89 & 86.41 & 74.70 & \textbf{84.41} & 82.29 & 83.86 & 0.00 & 0.00 & 77.26 & \textbf{77.45} \\
 & & & 1.0 & \textbf{91.35} & 86.95 & 78.6 & 81.27 & 45.21 & 78.20 & 75.57 & \textbf{78.55} & 0.00 & 0.00 & 70.31 & \textbf{71.29} \\
\midrule  
 & & \multirow{3}{*}{3} & 0.2 & \textbf{47.05} & 38.63 & 38.50 & / & \textbf{42.76} & 36.19 & 37.94 & / & 0.00 & 0.57 & \textbf{37.24} & / \\
 & & & 0.6 & \textbf{47.05} & 37.83 & 33.97 & / & 23.59 & \textbf{33.85} & 33.16 & / & 0.00 & 0.00 & \textbf{31.97} & / \\
 & & & 1.0 & \textbf{47.05} & 36.63 & 31.69 & / & 14.42 & \textbf{31.06} & 30.79 & / & 0.00 & 0.00 & \textbf{29.52} & / \\
\cmidrule(r){3-16}
 & & \multirow{3}{*}{5} & 0.2 & \textbf{47.05} & 38.73 & 39.75 & / & \textbf{41.54} & 35.97 & 38.86 & / & 0.00 & 0.19 & \textbf{37.77} & / \\
TinyImageNet & Box Blur & & 0.6 & \textbf{47.05} & 37.33 & 32.76 & / & 22.21 & \textbf{32.10} & 31.49 & / & 0.00 & 0.00 & \textbf{29.08} & / \\
 & & & 1.0 & \textbf{47.05} & 35.10 & 26.29 & / & 10.52 & \textbf{27.31} & 24.86 & / & 0.00 & 0.00 & \textbf{23.43} & / \\
\cmidrule(r){3-16}
 & & \multirow{3}{*}{7} & 0.2 & \textbf{47.05} & 39.09 & 40.01 & / & \textbf{41.36} & 36.31 & 39.10 & / & 0.00 & 0.08 & \textbf{37.73} & / \\
 & & & 0.6 & \textbf{47.05} & 36.00 & 31.41 & / & 22.46 & \textbf{30.49} & 29.50 & / & 0.00 & 0.00 & \textbf{26.54} & / \\
 & & & 1.0 & \textbf{47.05} & 28.80 & 23.36 & / & ~9.73 & 19.79 & \textbf{21.15} & / & 0.00 & 0.00 & \textbf{19.09} & / \\
\bottomrule
\end{tabular}
}
\vspace{0em}
\caption{Experimental evaluation on CNN7. $s$ is the perturbation kernel size,
$\epsilon$ the perturbation strength for training/verification. STD/PGD
denotes Standard and Adversarial Training, SSIP and RS denote Certified
Training based on SSIP and RSIP-SSIP bounding, respectively. Standard Acc.
is the accuracy on the test dataset, Empirical Acc. the empirical robust
accuracy, \ie the percentage of samples that are classified correctly under
a PGD adversarial attack. Robust Acc. is the verified robust accuracy
computed using RSIP-SSIP.}
\label{tab:experimental_evaluation}
\end{table*}
We built upon the bound propagation implementation of VeriNet
\cite{HenriksenLomuscio20,HenriksenLomuscio21}, extended it to handle
convolutional perturbations and added the required training functionality
as well as a PGD attack implementation.
PyTorch
\cite{Paszke+19} was used for vectorised computations and the efficient
training of models. In line with the SoA robust training literature, we
used the CNN7 architecture and the CIFAR10 as well as the TinyImageNet
datasets for our experiments. The experimental setup of using this
architecture and these datasets is well-accepted practice in certified
training works due to the high costs associated with the certified training
methods~\cite{Shi+21a,Palma+22,Mao+24} and the fact that bound computation
for newer architectures such as transformers is notoriously difficult due
to their nonlinearity~\cite{Shi+20}.
CNN7 consists of five convolutional and two fully-connected layers, all
layers except for the last one are followed by a ReLU activation layer.
\citet{Shi+21a} suggest multiple variants with and without batch
normalisation (BN) layers, we fully add BatchNorm layers since a
preliminary ablation study shows that this leads to better performance.
To demonstrate the
scalability of our method to larger model architectures, we present
additional results on a ResNet18 model.

CIFAR10 experiments on CNN7 were
run on an Ubuntu server with an AMD Ryzen Threadripper 3970X 32-Core CPU, a
Nvidia RTX 3090 GPU and 256GB of RAM. CIFAR10 experiments on ResNet18 and
TinyImageNet experiments used four cores of an Ubuntu server with an Intel
Xeon Platinum 8358 CPU, a Nvidia L40S GPU and 64GB of RAM.
\subsection{Experimental Setup}
\label{ssec:experimental_setup}
We tuned the number of warm-up and training epochs to
be used and found that the setting used by \citet{DePalma+24} with $80$
warm-up epochs during which the perturbation radius $\epsilon$ is linearly
increased up to its maximum followed by $80$ epochs of training with the
full perturbation magnitude yielded the best results. The longer schedule
with $80$ warm-up epochs and $260$ total epochs was outperformed by the
shorter $80/160$ schedule. In line with other robust training works we used
a learning rate of $10^{-5}$. We compared the learning rate schedule used
by other publications \cite{Shi+21a} which decays the learning rate twice
by a factor of $0.2$ after $120$ and $140$ epochs to a cosine annealing
learning rate schedule. Differences were small with the cosine annealing
schedule showing slightly increased standard accuracies at the expense of
slightly decreased robust accuracies due to its earlier learning rate
decay. We therefore used that schedule in our experiments to preserve
standard performance.\\
The results of our extensive evaluation are shown in Table
\ref{tab:experimental_evaluation}. We repeated our runs for $\epsilon \in
\{0.2, 0.6, 1.0\}$ and perturbation kernel sizes $s \in\{3, 5, 7\}$. For
each of these settings, we considered motion blur perturbations with a
blurring angle of $0^\circ$ as well as box blur and sharpen perturbations.
We further considered training setups combining either the PGD, SSIP or
RSIP-SSIP loss with a standard cross-entropy loss as outlined in Section
\ref{sec:method}. We evaluated each of the trained networks on the entire
test set and computed the standard accuracy, the empirical robust accuracy
using PGD and the verified robust accuracy using RSIP-SSIP. On
TinyImageNet \cite{DePalma+24,Mao+24} we used a slightly modified CNN7
\cite{Shi+21a} and restricted our analysis to box blur perturbations due to
training times. Here, RSIP-based training was infeasible due to the
substantial memory requirements associated with RSIP passes through
networks with large layers \cite{Mueller+20}. We therefore restricted our
experiments to PGD- and SSIP-based training with RSIP-SSIP only used for
test-time verification.

\subsection{Results on CIFAR10} \label{ssec:results_cifar10_discussion}
\paragraph{Standard (STD) and Adversarial Training (PGD)} Both methods
achieved high standard accuracies, for small perturbations the empirical
robustness of the models was close to the upper bound. For larger
perturbations only PGD models preserved high empirical robustness while
that of STD models dropped. However, the verified robustness was low since
these methods do not promote verifiability in any way, therefore we cannot
guarantee the non-existence of failure cases. This prevents the deployment
of such models in safety-critical domains.
\paragraph{SSIP and RSIP-SSIP Training}
The two robust training schemes produced models with significantly
increased robust accuracies. For small perturbations they preserved high
standard accuracies, for large perturbations (\eg motion and box blur at
$\epsilon=1.0, s=7$) the standard accuracies were reduced. On CIFAR10, the
more precise RSIP-SSIP bounding induced less implicit regularisation due to
tighter bounds, leading to slightly lower verified robustness but higher
standard accuracies. For very large perturbations, RSIP-SSIP on CIFAR10
achieved a higher robust accuracy than the SSIP-trained models due to
the heavily reduced standard accuracy of the SSIP models. On TinyImageNet,
SSIP provided robustified networks with small drops in standard accuracy
for moderate perturbation sizes and significantly higher robust accuracies.
For large perturbations, \eg box blur perturbations with kernel size $s=7$
and perturbation size $\eps=1.0$ we still attained a verified robustness
of $19.09\%$, albeit with a larger standard performance decrease due to the
required strong training-time regularisation.\\

\begin{table*}[!htb]
\small
\centering
\adjustbox{max width=.75\textwidth}{%
\begin{tabular}{cccccccccccc} \toprule
\multicolumn{3}{c}{} & \multicolumn{3}{c}{Standard Acc.} & \multicolumn{3}{c}{Empirical Acc.} & \multicolumn{3}{c}{Robust Acc.} \\
\cmidrule(r){4-6} \cmidrule(r){7-9} \cmidrule(r){10-12}
Perturbation & $s$ & $\epsilon$ & STD & PGD & SSIP & STD & PGD & SSIP & STD & PGD & SSIP \\
\cmidrule(lr){1-3} \cmidrule(lr){4-12}
\multirow{9}{*}{Box Blur} & \multirow{3}{*}{3} & 0.2 & \textbf{94.48} & 91.08 & 87.70 & \textbf{92.79} & 90.16 & 86.46 & 0.00 & 0.00 & \textbf{84.70} \\
 &  & 0.6 & \textbf{94.48} & 90.44 & 81.64 & 84.87 & \textbf{87.82} & 79.80 & 0.00 & 0.00 & \textbf{75.42} \\
 &  & 1.0 & \textbf{94.48} & 90.46 & 77.22 & 62.02 & \textbf{84.88} & 74.84 & 0.00 & 0.00 & \textbf{70.48} \\
\cmidrule(r){2-12}
 & \multirow{3}{*}{5} & 0.2 & \textbf{94.48} & 90.84 & 88.58 & \textbf{92.77} & 90.18 & 87.74 & 0.00 & 0.00 & \textbf{85.26} \\
 &  & 0.6 & \textbf{94.48} & 91.20 & 79.68 & 83.32 & \textbf{88.20} & 75.82 & 0.00 & 0.00 & \textbf{66.40} \\
 &  & 1.0 & \textbf{94.48} & 89.90 & 67.92 & 46.31 & \textbf{81.44} & 64.02 & 0.00 & 0.00 & \textbf{58.88} \\
\cmidrule(r){2-12}
 & \multirow{3}{*}{7} & 0.2 & \textbf{94.48} & 90.98 & 87.92 & \textbf{92.64} & 90.06 & 87.18 & 0.00 & 0.00 & \textbf{84.42} \\
 &  & 0.6 & \textbf{94.48} & 89.98 & 81.16 & 84.44 & \textbf{86.86} & 77.98 & 0.00 & 0.00 & \textbf{68.68} \\
 &  & 1.0 & \textbf{94.48} & 88.58 & 66.64 & 45.68 & \textbf{75.40} & 60.18 & 0.00 & 0.00 & \textbf{55.46} \\
\midrule
\multirow{9}{*}{Sharpen} & \multirow{3}{*}{3} & 0.2 & \textbf{94.48} & 92.14 & 88.28 & \textbf{92.87} & 91.66 & 87.98 & 0.00 & 0.00 & \textbf{86.04} \\
 &  & 0.6 & \textbf{94.48} & 92.20 & 85.24 & \textbf{91.24} & 91.10 & 84.02 & 0.00 & 0.00 & \textbf{80.62} \\
 &  & 1.0 & \textbf{94.48} & 91.94 & 82.96 & 89.23 & \textbf{90.90} & 81.68 & 0.00 & 0.00 & \textbf{76.08} \\
\cmidrule(r){2-12}
 & \multirow{3}{*}{5} & 0.2 & \textbf{94.48} & 91.60 & 88.24 & \textbf{92.84} & 90.98 & 87.72 & 0.00 & 0.00 & \textbf{86.28} \\
 &  & 0.6 & \textbf{94.48} & 91.84 & 84.80 & 90.86 & \textbf{90.96} & 83.12 & 0.00 & 0.00 & \textbf{78.04} \\
 &  & 1.0 & \textbf{94.48} & 92.12 & 81.32 & 88.46 & \textbf{91.10} & 79.32 & 0.00 & 0.00 & \textbf{71.48} \\
\cmidrule(r){2-12}
 & \multirow{3}{*}{7} & 0.2 & \textbf{94.48} & 92.02 & 88.92 & \textbf{92.81} & 91.48 & 88.14 & 0.00 & 0.00 & \textbf{85.50} \\
 &  & 0.6 & \textbf{94.48} & 92.10 & 85.78 & 90.77 & \textbf{90.92} & 84.34 & 0.00 & 0.00 & \textbf{79.44} \\
 &  & 1.0 & \textbf{94.48} & 92.42 & 83.30 & 88.37 & \textbf{91.08} & 81.16 & 0.00 & 0.00 & \textbf{73.68} \\
\midrule
\multirow{9}{*}{Motion Blur} & \multirow{3}{*}{3} & 0.2 & \textbf{94.48} & 91.66 & 87.78 & \textbf{93.06} & 90.90 & 87.18 & 0.00 & 0.00 & \textbf{86.10} \\
 &  & 0.6 & \textbf{94.48} & 90.94 & 85.36 & \textbf{89.22} & 88.82 & 84.62 & 0.00 & 0.00 & \textbf{82.86} \\
 &  & 1.0 & \textbf{94.48} & 90.88 & 83.82 & 82.69 & \textbf{88.30} & 82.48 & 0.00 & 0.00 & \textbf{80.36} \\
\cmidrule(r){2-12}
 & \multirow{3}{*}{5} & 0.2 & \textbf{94.48} & 91.10 & 87.56 & \textbf{92.72} & 90.10 & 86.86 & 0.00 & 0.00 & \textbf{84.70} \\
 &  & 0.6 & \textbf{94.48} & 91.22 & 80.86 & 85.07 & \textbf{88.76} & 79.24 & 0.00 & 0.00 & \textbf{76.72} \\
 &  & 1.0 & \textbf{94.48} & 90.58 & 79.02 & 63.36 & \textbf{85.62} & 76.94 & 0.00 & 0.00 & \textbf{73.02} \\
\cmidrule(r){2-12}
 & \multirow{3}{*}{7} & 0.2 & \textbf{94.48} & 91.08 & 87.28 & \textbf{92.62} & 89.88 & 86.32 & 0.00 & 0.00 & \textbf{83.64} \\
 &  & 0.6 & \textbf{94.48} & 91.00 & 78.00 & 82.85 & \textbf{87.88} & 75.64 & 0.00 & 0.00 & \textbf{70.76} \\
 &  & 1.0 & \textbf{94.48} & 89.16 & 74.76 & 51.19 & \textbf{81.30} & 71.78 & 0.00 & 0.00 & \textbf{67.66} \\
\bottomrule
\end{tabular}
}
\vspace{0em}
\caption{Experimental evaluation on ResNet18 using the CIFAR10 dataset. $s$
is the perturbation kernel size, $\epsilon$ the perturbation strength for
training/verification. STD/PGD denotes Standard and Adversarial Training,
SSIP denotes Certified Training based on SSIP bounding. Standard Acc. is
the accuracy on the test dataset, Empirical Acc. the empirical robust
accuracy, \ie the percentage of samples that are classified correctly under
a PGD adversarial attack. Robust Acc. is the verified robust accuracy
computed using RSIP-SSIP.}
\label{tab:experimental_evaluation_resnet18}
\end{table*}

\noindent \textbf{Box Blur} perturbations introduce substantial corruptions
to images~\cite{BruecknerLomuscio25}, so robustification against them lead
to the largest drop in standard performance. This is consistent with the
well-known trade-off between accuracy and robustness \cite{Zhang+19b}.
Nevertheless, on CIFAR10, we still obtained robust accuracies above
$70\%$ in all but two cases. On TinyImageNet, we observed similar trends
with robustification for smaller perturbation strengths achieving high
standard and robust accuracies while for larger perturbation strengths
the tradeoff between accuracy and robustness is more evident.\\
\textbf{Motion Blur} perturbations have much milder effects on inputs,
allowing for improved accuracy-robustness tradeoffs. For reasonable
perturbation sizes on CIFAR10 ($s=3, \epsilon \in \{0.2, 0.6, 1.0\}$) our
method increased the robust accuracy from PGD's $0\%$ to up to $87.92\%$
while reducing the standard accuracy by at most $1\%$. For large
perturbation strengths, robust accuracies remained above $70\%$ but lead
to more evident drops in standard accuracy.\\
\textbf{Sharpen} perturbations only introduce small changes to inputs.
Across all perturbation strengths, certified training achieved strong
robust accuracies above $80\%$ while trading off at most $3\%$ in
standard performance compared to Standard Training in the worst case.

\paragraph{Summary} Using CNN7 on CIFAR10, the state-of-the-art method for
Certified Training on white noise achieves a standard accuracy of just
$80.61\%$ for a relatively small $\epsilon=\nicefrac{2}{255}$ with a robust
accuracy of $61.65\%$ \cite{DePalma+24}. This is well below the accuracies
for many of our scenarios. Although robustification against white noise is
a more difficult problem, our method is very effective at navigating the
trade-off between accuracy and robustness.
While our method trains models with formal robustness guarantees,
no such guarantees can be provided for models trained via STD or PGD.

\subsection{Results on ResNet18}
\label{ssec:results_on_resnet18}
We conduct standard, adversarial and SSIP-based robust training runs
on a ResNet18 model using the CIFAR10 dataset. Due to the size of the
model, RSIP-based robust training leads to substantial memory requirements
and is therefore infeasible here. The results of our evaluation are shown
in Table~\ref{tab:experimental_evaluation_resnet18}. We generally observe
trends that are similar to those for CNN7. However, we find that ResNet18
models achieves higher standard accuracies on the dataset compared to CNN7
due to its increased model capacity. We also observe that, without robust
training, the robust accuracy from single bound propagation passes is zero
for both standard and adversarial training across all settings. For models
with a depth similar to that of a ResNet18 model, robust training is
necessary to achieve a nonzero verified robustness using single bound
propagation passes. Comparing the performance of the ResNet18 models with
that of CNN7, we observe trends in line with those found in prior
works~\cite{Mao+24b}: Due to the large depth of the ResNet model, robust
training introduces substantial regularisation at training time in order to
preserve verifiability of the neural network. While this leads to increased
robustness compared to models trained with standard training, it comes at a
high cost in terms of standard accuracy. For the robustly trained models,
we generally observe lower standard accuracies for ResNet18 compared to
CNN7 due to the heavy regularisation. Nevertheless, we are able to
demonstrate that our robust training method extends to larger network
architectures and that it can robustify those to a significant extent. We
find that, for ResNet18, correctly choosing the weight decay correctly
plays an important role and leads to differing accuracies. To account for
this, we tune the weight decay for each training run by conducting a run
for values of $10^{-3}, 5 \times 10^{-4}, 10^{-4}$ and selecting the run
with the highest standard accuracy.

Besides weight decay, we find that the hyperparameters we used for CNN7
work well for ResNet18 and also add batch normalisation layers to the
ResNet18 model.

\section{Conclusions}
\label{sec:conclusions}

We present the first method for training models that are both accurate and
verifiably robust against convolutional perturbations. Robustness can be
significantly increased compared to models trained with Adversarial
Training which is useful for deploying models in a range of real-life
situations where motion blur or optic-created blurring can occur. The
verification-based approach evaluates the trained models in a dense
neighbourhood of inputs, providing formal safety guarantees before
deployment. For reasonable perturbation strengths, our most precise robust
training approach improves verified accuracy from $0\%$ to $82\%$ for
motion blur on CIFAR10 while only suffering from a $1\%$ decrease in
accuracy.
We also find that our method can equally be applied to much deeper ResNet18
models which illustrates the scalability of our approach.

\section*{Acknowledgements}
Benedikt Br\"uckner acknowledges support from the UKRI Centre for Doctoral Training in Safe and Trusted Artificial Intelligence [EP/S023356/1]. Alessio Lomuscio acknowledges partial support from the Royal Academy of Engineering via a Chair of Emerging Technologies.

{
    \small
    \bibliographystyle{ieeenat_fullname}
    \bibliography{bib}
}

\clearpage
\setcounter{page}{1}
\setcounter{section}{0}
\renewcommand{\thesection}{\Alph{section}}
\maketitlesupplementary
\section{Further Details on our Method}
\subsection{Motion Blur Example}
\label{ssec:motion_blur_example}
To illustrate the perturbation encoding in the layers that are prepended to
a neural network, we provide an example of a motion blur perturbation being
applied to a small image in the following:
\begin{example} \label{ex:perturbation_encoding}
For the sample calculations assume that an image $\mJ
\in \sR^{4 \times 4}$ and a motion blur kernel $\mM \in
\sR^{3 \times 3}$ simulating blurring along the angle $\phi = 0^\circ$ are
given as
\begin{equation}
\mM = \begin{pmatrix}
0 & \frac{1}{3} & 0 \\[2pt]
0 & \frac{1}{3} & 0 \\[2pt]
0 & \frac{1}{3} & 0
\end{pmatrix}, \quad
\mJ = \begin{pmatrix}
0 & 0 & 0 & 0 \\
0 & 3 & 6 & 0 \\
0 & 3 & 6 & 0 \\
0 & 0 & 0 & 0 \\
\end{pmatrix}
\end{equation}
For the sake of simplicity we allow pixel values outside $[0, 1]$ for this
example only. Using the information that the parameterised kernel $\mG_z$
must be equal to the identity kernel for $z=0$ and equal to $\mM$ for
$z=1$, the parameterisation is computed as
\begin{equation}
\mM_z = 
\underbrace{\begin{pmatrix}
0 & \frac{1}{3} & 0 \\[2pt]
0 & -\frac{2}{3} & 0 \\[2pt]
0 & \frac{1}{3} & 0
\end{pmatrix}}_{\mA}
z +
\underbrace{\begin{pmatrix}
0 & 0 & 0 \\[2pt]
0 & 1 & 0 \\[2pt]
0 & 0 & 0 \\
\end{pmatrix}}_{\mB}
\end{equation}
The result of the convolution of the batched image with the coefficient and
bias kernel is
\begin{align}
\mR_\mA &= \begin{pmatrix}
0 & 2 & 4 & 0 \\
0 & -1 & -2 & 0 \\
0 & -1 & -2 & 0 \\
0 & 2 & 4 & 0 \\
\end{pmatrix}\\
\mR_\mB &= \begin{pmatrix}
0 & 0 & 0 & 0 \\
0 & 3 & 6 & 0 \\
0 & 3 & 6 & 0 \\
0 & 0 & 0 & 0 \\
\end{pmatrix}
\end{align}
As an example, for $z=0$ it is obvious that
\begin{equation}
\mJ_\text{perturbed}  = \mR_\mA \cdot 0 + \mR_\mB = \mR_\mB = \mJ
\end{equation}
To obtain a slightly blurred image, set $z=0.5$ and calculate the result as
\begin{align}
\mJ_\text{perturbed}  &= \mR_\mA \cdot 0.5 + \mR_\mB\\
&= \begin{pmatrix}
0 & 1 & 2 & 0 \\
0 & 2.5 & 5 & 0 \\
0 & 2.5 & 5 & 0 \\
0 & 1 & 2 & 0 \\
\end{pmatrix}
\end{align}
One can see that the result is an image which has been blurred along the
$\phi=0^\circ$ axis.
\end{example}
\subsection{Perturbation Batching Example}
\label{ssec:perturbation_batching_example}
To demonstrate the batched operations computed by our custom perturbation
encoding, we provide some sample calculations detailing the maths behind
them in the following.
\begin{example} \label{ex:batched_operation}
Assume that two result tensors $\mR_\mA, \mR_\mB \in \sR^{2 \times 1
\times 2 \times 2}$ are given as
\begin{align}
\mR_\mA[1, 1, :, :] = \begin{pmatrix}
0 & 3 \\
2 & 1 \\
\end{pmatrix}, 
\mR_\mA[2, 1, :, :] = \begin{pmatrix}
1 & 0 \\
3 & 2 \\
\end{pmatrix}\\
\mR_\mB[1, 1, :, :] = \begin{pmatrix}
1 & 1 \\
1 & 1 \\
\end{pmatrix}, 
\mR_\mB[2, 1, :, :] = \begin{pmatrix}
2 & 2 \\
2 & 2 \\
\end{pmatrix}
\end{align}
We use small tensors here due to space constraints. However, all of the
following computations can equally be run for larger matrices such as the
ones from Example \ref{ssec:motion_blur_example}. In line with equation
\ref{eq:tensor_transform_1} and \ref{eq:tensor_transform_2} we calculate
\begin{align}
\mR_\mA[:, :, 1]' = \begin{pmatrix}
0 & 3 & 2 & 1 \\
1 & 0 & 3 & 2 \\
\end{pmatrix},\\
\mR_\mB[:, :, 1]' = \begin{pmatrix}
1 & 1 & 1 & 1 \\
2 & 2 & 2 & 2 \\
\end{pmatrix},
\end{align}
Choosing $\tilde{\vz} = \begin{pmatrix}
0.5 & 1
\end{pmatrix}^T$
as the perturbation vector, the batched matrix multiplication
(see equation \ref{eq:batched_operation}) returns
\begin{align}
\mI_\text{perturbed}' &= 
\begin{pmatrix}
0 & 1.5 & 1 & 0.5 \\
1 & 0 & 3 & 2 \\
\end{pmatrix} + 
\begin{pmatrix}
1 & 1 & 1 & 1 \\
2 & 2 & 2 & 2 \\
\end{pmatrix} \\
&=
\begin{pmatrix}
1 & 2.5 & 2 & 1.5 \\
3 & 2 & 5 & 4 \\
\end{pmatrix}
\end{align}
Using Equation \ref{eq:tensor_transform_3}, the final output of the
operation in the correct shape is $\mI_\text{perturbed}$ where
\begin{align}
\mI_\text{perturbed}[1, 1, :, :] &= \begin{pmatrix}
1 & 2.5 \\
2 & 1.5 \\
\end{pmatrix}\\
\mI_\text{perturbed}[2, 1, :, :] &= \begin{pmatrix}
3 & 2 \\
5 & 4 \\
\end{pmatrix}
\end{align}
\end{example}
\subsection{Perturbation Kernels Used in This Work}
\label{sec:perturbation_kernels}

Robust Training is performed for box blur, sharpen and motion blur
perturbations with perturbation kernels in different sizes $s$. The details
of the parameterisations that are used in defining those kernels are
provided in the following sections. All of the parameterisations are
defined as $\mK = \mA \cdot \vz + \mB$ with $\mK, \mA, \mB \in
\mathbb{R}^{s \times s}$ \cite{BruecknerLomuscio25}. By inspecting the
$z=0$ case of this parameterisation, it is clear that for a kernel $\mK$ of
size $s$, it always holds that $\mB$, i.e. the bias matrix, is the
$s$-dimensional identity matrix. In the following we therefore only present
the coefficient matrices $\mA$ we consider.
\subsection{Box Blur}
\paragraph{Kernel Size $s=3$}
\begin{equation}
\scriptstyle
\mA = \begin{pmatrix}
0.11 & 0.11 & 0.11 \\
0.11 & -0.89 & 0.11 \\
0.11 & 0.11 & 0.11 \\
\end{pmatrix}\\
\end{equation}
\paragraph{Kernel Size $s=5$}
\begin{equation}
\mA = \begin{pmatrix}
0.04 & 0.04 & 0.04 & 0.04 & 0.04 \\
0.04 & 0.04 & 0.04 & 0.04 & 0.04 \\
0.04 & 0.04 & -0.96 & 0.04 & 0.04 \\
0.04 & 0.04 & 0.04 & 0.04 & 0.04 \\
0.04 & 0.04 & 0.04 & 0.04 & 0.04 \\
\end{pmatrix}\\
\end{equation}
\paragraph{Kernel Size $s=7$}
\begin{equation}
\resizebox{0.8\hsize}{!}{
$\mA = \begin{pmatrix}
0.02 & 0.02 & 0.02 & 0.02 & 0.02 & 0.02 & 0.02 \\
0.02 & 0.02 & 0.02 & 0.02 & 0.02 & 0.02 & 0.02 \\
0.02 & 0.02 & 0.02 & 0.02 & 0.02 & 0.02 & 0.02 \\
0.02 & 0.02 & 0.02 & -0.98 & 0.02 & 0.02 & 0.02 \\
0.02 & 0.02 & 0.02 & 0.02 & 0.02 & 0.02 & 0.02 \\
0.02 & 0.02 & 0.02 & 0.02 & 0.02 & 0.02 & 0.02 \\
0.02 & 0.02 & 0.02 & 0.02 & 0.02 & 0.02 & 0.02 \\
\end{pmatrix}$
}\\
\end{equation}
\subsection{Sharpen}
\paragraph{Kernel Size $s=3$}
\begin{equation}
\mA = \begin{pmatrix}
0.00 & -0.25 & 0.00 \\
-0.25 & 1.00 & -0.25 \\
0.00 & -0.25 & 0.00 \\
\end{pmatrix}\\
\end{equation}
\paragraph{Kernel Size $s=5$}
\begin{equation}
\mA = \begin{pmatrix}
0.00 & 0.00 & -0.08 & 0.00 & 0.00 \\
0.00 & -0.08 & -0.08 & -0.08 & 0.00 \\
-0.08 & -0.08 & 1.00 & -0.08 & -0.08 \\
0.00 & -0.08 & -0.08 & -0.08 & 0.00 \\
0.00 & 0.00 & -0.08 & 0.00 & 0.00 \\
\end{pmatrix}\\
\end{equation}
\paragraph{Kernel Size $s=7$}
\begin{equation}
\resizebox{0.8\hsize}{!}{
$\mA = \begin{pmatrix}
0.00 & 0.00 & 0.00 & -0.04 & 0.00 & 0.00 & 0.00 \\
0.00 & 0.00 & -0.04 & -0.04 & -0.04 & 0.00 & 0.00 \\
0.00 & -0.04 & -0.04 & -0.04 & -0.04 & -0.04 & 0.00 \\
-0.04 & -0.04 & -0.04 & 1.00 & -0.04 & -0.04 & -0.04 \\
0.00 & -0.04 & -0.04 & -0.04 & -0.04 & -0.04 & 0.00 \\
0.00 & 0.00 & -0.04 & -0.04 & -0.04 & 0.00 & 0.00 \\
0.00 & 0.00 & 0.00 & -0.04 & 0.00 & 0.00 & 0.00 \\
\end{pmatrix}$
}\\
\end{equation}
\subsection{Motion Blur, $\phi = 0^\circ$}
\paragraph{Kernel Size $s=3$}
\begin{equation}
\mA = \begin{pmatrix}
0.00 & 0.33 & 0.00 \\
0.00 & -0.67 & 0.00 \\
0.00 & 0.33 & 0.00 \\
\end{pmatrix}\\
\end{equation}
\paragraph{Kernel Size $s=5$}
\begin{equation}
\mA = \begin{pmatrix}
0.00 & 0.00 & 0.20 & 0.00 & 0.00 \\
0.00 & 0.00 & 0.20 & 0.00 & 0.00 \\
0.00 & 0.00 & -0.80 & 0.00 & 0.00 \\
0.00 & 0.00 & 0.20 & 0.00 & 0.00 \\
0.00 & 0.00 & 0.20 & 0.00 & 0.00 \\
\end{pmatrix}\\
\end{equation}
\paragraph{Kernel Size $s=7$}
\begin{equation}
\resizebox{0.8\hsize}{!}{
$\mA = \begin{pmatrix}
0.00 & 0.00 & 0.00 & 0.14 & 0.00 & 0.00 & 0.00 \\
0.00 & 0.00 & 0.00 & 0.14 & 0.00 & 0.00 & 0.00 \\
0.00 & 0.00 & 0.00 & 0.14 & 0.00 & 0.00 & 0.00 \\
0.00 & 0.00 & 0.00 & -0.86 & 0.00 & 0.00 & 0.00 \\
0.00 & 0.00 & 0.00 & 0.14 & 0.00 & 0.00 & 0.00 \\
0.00 & 0.00 & 0.00 & 0.14 & 0.00 & 0.00 & 0.00 \\
0.00 & 0.00 & 0.00 & 0.14 & 0.00 & 0.00 & 0.00 \\
\end{pmatrix}$
}\\
\end{equation}
\section{Further Details on the Experimental Setup and Results}
\label{sec:further_details}
We use the CIFAR10 dataset \cite{KrizhevskyNairHinton14} which contains
colour images of size $32 \times 32$ for our experimental evaluation.
Images belong to one of ten different classes such as automobile, airplane
or bird which are mutually exclusive. We conduct our training runs on the
training dataset which contains $50,000$ and then run our evaluation on the
test dataset with its $10,000$ test images. We calculate the mean and
standard deviation of each channel on the training dataset and use those
values to normalise both the training and test set images. To improve
performance, we further employ random horizontal flipping and random
cropping as done by other works in the field \cite{DePalma+24}. The
architecture of the employed CNN7 network is shown in Table \ref{tab:cnn7}.
Additional details on the parameters used for our training runs are
provided in Table \ref{tab:additional_parameters}. The weight decay
parameter was chosen after evaluating $10^{-3}, 5 \times 10^{-4}, 10^{-4}$
for it and obtaining the best results in terms of standard accuracy for the
value of $5 \times 10^{-4}$. We further use TinyImageNet, a downsampled and
smaller version of ImageNet, for our experiments. This dataset contains $64
\times 64$ RGB images from $200$ classes with the training set containing
$100,000$ samples while the test set contains a further $10,000$ images. We
similarly use normalisation and random horizontal flips and random cropping
here as is standard practice in the literature. In line with usual
practice, we further evaluate networks trained on TinyImageNet on the
validation set.
\begin{table}[htb]
\centering
\small
\begin{tabular}{ccc}
\toprule
Index & Layer Type & Parameters \\
\midrule
1 & Convolutional & \makecell{Width: $64$ \\ Kernel Size: $3$ \\ Stride: $1$ \\ Padding: $1$} \\
\midrule
2 & BatchNorm + ReLU & - \\
\midrule
3 & Convolutional & \makecell{Width: $64$ \\ Kernel Size: $3$ \\ Stride: $1$ \\ Padding: $1$} \\
\midrule
4 & BatchNorm + ReLU & - \\
\midrule
5 & Convolutional & \makecell{Width: $128$ \\ Kernel Size: $3$ \\ Stride: $2$ \\ Padding: $1$} \\
\midrule
6 & BatchNorm + ReLU & - \\
\midrule
7 & Convolutional & \makecell{Width: $128$ \\ Kernel Size: $3$ \\ Stride: $1$ \\ Padding: $1$} \\
\midrule
8 & BatchNorm + ReLU & - \\
\midrule
9 & Convolutional & \makecell{Width: $128$ \\ Kernel Size: $3$ \\ Stride: $1$ or $2^\dagger$ \\ Padding: $1$} \\
\midrule
10 & BatchNorm + ReLU & - \\
\midrule
11 & Linear & No. of Neurons: $512$ \\
\midrule
12 & BatchNorm + ReLU & - \\
\midrule
13 & Linear & No. of Neurons: $10$ \\
\bottomrule
\end{tabular}
\caption{The CNN7 network used throughout this work. ${}^\dagger$ A stride of $1$ is used for the architecture employed on CIFAR10 while a stride of $2$ is used for TinyImageNet.}
\label{tab:cnn7}
\end{table}
\begin{table}[htb]
\centering
\begin{tabular}{cc}
\toprule
Parameter & Value \\
\midrule
Warm-up epochs & $80$ \\
Total epochs & $160$ \\
Learning rate & $1 \times 10^{-5}$ \\
Learning rate scheduler & Cosine Annealing \\
Weight decay & $5 \times 10^{-4}$ \\
Batch size & $128$ \\
Number of PGD Steps & $8$ \\
PGD step size & $0.25$ \\
\bottomrule
\end{tabular}
\caption{Additional parameters used throughout our training runs}
\label{tab:additional_parameters}
\end{table}
\begin{table}[htb]
\small
\centering
\begin{tabular}{cccc}
\toprule
Model & Dataset & Method & Average Runtime\\
\midrule
\multirow{5}[6]{*}{CNN7} & & PGD & $6.50 \times 10^3$\\
 & CIFAR10 & SSIP & $2.29 \times 10^4$\\
 & & RSIP-SSIP & $3.96 \times 10^4$\\
\cmidrule(r){2-4}
 & \multirow{2}{*}{TinyImageNet} & PGD & $4.84 \times 10^4$\\
 & & SSIP & $1.10 \times 10^5$\\
\midrule
\multirow{2}{*}{ResNet18} & \multirow{2}{*}{CIFAR10} & PGD & $1.25 \times 10^{4}$\\
 & & SSIP & $4.37 \times 10^4$\\
\bottomrule
\end{tabular}
\caption{Average runtimes for CNN7 and ResNet18 training on CIFAR10 and
TinyImageNet. PGD denotes Adversarial Training, SSIP and RSIP-SSIP denote
Certified Training based on the respective bounding method.}
\label{tab:runtimes_cifar10}
\end{table}
We also report the average time required for the training runs presented in
Section \ref{sec:evaluation} in Table \ref{tab:runtimes_cifar10}. As
expected and as is known in the literature, Certified Training with either
SSIP or RSIP-SSIP requires additional time when compared to pure
Adversarial Training. This is due to the cost of the bound computation
which is incurred for every training batch. However, our experiments prove
that, depending on the scenario at hand, this additional cost can be
justified due to the superior robustness that networks trained using certified training
exhibit compared to the adversarially trained ones.

\begin{table}[!htb]
\small
\centering
\adjustbox{max width=\textwidth}{%
\begin{tabular}{cccccc} \toprule
Perturbation & $s$ & $\epsilon$ & STD & PGD & SSIP \\
\cmidrule(lr){1-3} \cmidrule(lr){4-6}
\multirow{9}{*}{Box Blur} & \multirow{3}{*}{3} & 0.2 & $1 \times 10^{-4}$ & $1 \times 10^{-4}$ & $1 \times 10^{-3}$ \\
 &  & 0.6 & $1 \times 10^{-4}$ & $1 \times 10^{-4}$ & $1 \times 10^{-3}$ \\
 &  & 1.0 & $1 \times 10^{-4}$ & $5 \times 10^{-4}$ & $1 \times 10^{-4}$ \\
\cmidrule(r){2-6}
 & \multirow{3}{*}{5} & 0.2 & $1 \times 10^{-4}$ & $1 \times 10^{-4}$ & $1 \times 10^{-4}$ \\
 &  & 0.6 & $1 \times 10^{-4}$ & $1 \times 10^{-4}$ & $5 \times 10^{-4}$ \\
 &  & 1.0 & $1 \times 10^{-4}$ & $5 \times 10^{-4}$ & $5 \times 10^{-4}$ \\
\cmidrule(r){2-6}
 & \multirow{3}{*}{7} & 0.2 & $1 \times 10^{-4}$ & $1 \times 10^{-4}$ & $5 \times 10^{-4}$ \\
 &  & 0.6 & $1 \times 10^{-4}$ & $5 \times 10^{-4}$ & $5 \times 10^{-4}$ \\
 &  & 1.0 & $1 \times 10^{-4}$ & $5 \times 10^{-4}$ & $5 \times 10^{-4}$ \\
\midrule
\multirow{9}{*}{Sharpen} & \multirow{3}{*}{3} & 0.2 & $1 \times 10^{-4}$ & $1 \times 10^{-4}$ & $1 \times 10^{-4}$ \\
 &  & 0.6 & $1 \times 10^{-4}$ & $5 \times 10^{-4}$ & $1 \times 10^{-3}$ \\
 &  & 1.0 & $1 \times 10^{-4}$ & $1 \times 10^{-4}$ & $1 \times 10^{-3}$ \\
\cmidrule(r){2-6}
 & \multirow{3}{*}{5} & 0.2 & $1 \times 10^{-4}$ & $1 \times 10^{-4}$ & $1 \times 10^{-4}$ \\
 &  & 0.6 & $1 \times 10^{-4}$ & $1 \times 10^{-4}$ & $5 \times 10^{-4}$ \\
 &  & 1.0 & $1 \times 10^{-4}$ & $1 \times 10^{-4}$ & $5 \times 10^{-4}$ \\
\cmidrule(r){2-6}
 & \multirow{3}{*}{7} & 0.2 & $1 \times 10^{-4}$ & $1 \times 10^{-4}$ & $5 \times 10^{-4}$ \\
 &  & 0.6 & $1 \times 10^{-4}$ & $1 \times 10^{-4}$ & $5 \times 10^{-4}$ \\
 &  & 1.0 & $1 \times 10^{-4}$ & $1 \times 10^{-4}$ & $5 \times 10^{-4}$ \\
\midrule
\multirow{9}{*}{Motion Blur} & \multirow{3}{*}{3} & 0.2 & $1 \times 10^{-4}$ & $1 \times 10^{-4}$ & $1 \times 10^{-3}$ \\
 &  & 0.6 & $1 \times 10^{-4}$ & $5 \times 10^{-4}$ & $1 \times 10^{-4}$ \\
 &  & 1.0 & $1 \times 10^{-4}$ & $1 \times 10^{-4}$ & $5 \times 10^{-4}$ \\
\cmidrule(r){2-6}
 & \multirow{3}{*}{5} & 0.2 & $1 \times 10^{-4}$ & $5 \times 10^{-4}$ & $1 \times 10^{-4}$ \\
 &  & 0.6 & $1 \times 10^{-4}$ & $1 \times 10^{-4}$ & $1 \times 10^{-3}$ \\
 &  & 1.0 & $1 \times 10^{-4}$ & $1 \times 10^{-4}$ & $5 \times 10^{-4}$ \\
\cmidrule(r){2-6}
 & \multirow{3}{*}{7} & 0.2 & $1 \times 10^{-4}$ & $5 \times 10^{-4}$ & $1 \times 10^{-4}$ \\
 &  & 0.6 & $1 \times 10^{-4}$ & $1 \times 10^{-4}$ & $1 \times 10^{-3}$ \\
 &  & 1.0 & $1 \times 10^{-4}$ & $5 \times 10^{-4}$ & $1 \times 10^{-4}$ \\
\bottomrule
\end{tabular}
}
\vspace{0em}
\caption{Weight decay values used for the best-performing run on ResNet18}
\label{tab:weight_decay_resnet18}
\end{table}

\section{Ablation Studies on CIFAR10} \label{sec:ablation_studies}
\subsection{Ablation: Batch Normalisation} \label{ssec:ablation_batch_norm}
To check whether fully adding Batch Normalisation layers to the CNN7
architecture as suggested for IBP-based Certified Training by
\citet{Shi+21a} is beneficial for training neural networks robust to
convolutional perturbations, we conduct a small initial ablation study on CIFAR10. Box
Blur perturbations with a kernel size of $s=3$ and a strength of
$\epsilon=0.2$ are considered as the noise model for this evaluation. We
conduct training runs for multiple lengths of training schedules with the
CNN7 network as shown in Table \ref{tab:cnn7}, but also test CNN7 with all
batch normalisation layers removed and refer to it as CNN7*. We train the
network using SSIP robust training and evaluate the standard as well as the
robust accuracy computed by the SSIP bounds. The results are shown in Table
\ref{tab:ablation_batch_norm}. The $80/160$ schedule outperforms all other
schedules, although the differences in performance are small compared to
some of them. We also investigate the benefit of longer training schedules
again for stronger perturbations with $\epsilon=1.0$. However, for this
separate experiment, we also find that the $80/160$ performs best.\\
Comparing the performance of CNN7 and CNN7*, we find that CNN7 achieves
strictly higher standard and robust accuracies than CNN7*. The differences
in performance for the two models are already significant in terms of
standard accuracy where we see an improvement in performance of at least
$1\%$ of CNN7 compared to CNN7*. In terms of robust accuracy, the advantage
of CNN7 over CNN7* is even more pronounced with CNN7 reaching robust
accuracies up to $3.5\%$ higher than those for the model without batch
normalisation. We conclude that batch normalisation layers are not only
useful for Certified Training using IBP bounds, but also beneficial in our
case where we use tighter symbolic bounds to construct our training loss.
\begin{table}[htb]
\centering
\begin{tabular}{cccc}
\toprule
\makecell{Warm-up/Total\\epochs} & Model & Standard Acc. & Robust Acc. \\
\midrule
\multirow{2}{*}{$40/80$} & CNN7 & 86.80 & 79.44 \\
& CNN7* & 88.72 & 83.08 \\
\midrule
\multirow{2}{*}{$60/120$} & CNN7 & 87.28 & 80.60 \\
& CNN7* & 89.46 & 84.18 \\
\midrule
\multirow{2}{*}{$80/160$} & CNN7 & 88.00 & 81.5 \\
& CNN7* & 89.68 & 84.24 \\
\midrule
\multirow{2}{*}{$100/200$} & CNN7 & 87.06 & 80.68 \\
& CNN7* & 89.40 & 84.16 \\
\midrule
\multirow{2}{*}{$120/240$} & CNN7 & 87.64 & 80.54 \\
& CNN7* & 88.62 & 82.92 \\
\bottomrule
\end{tabular}
\caption{Ablation Study on using CNN7 with versus without batch
normalisation layers on CIFAR10. Networks are trained using SSIP robust training on
box blur perturbations with $s=3, \epsilon = 0.2$}
\label{tab:ablation_batch_norm}
\end{table}

\subsection{Ablation: IBP Initialisation and Warm-Up Loss}
\label{ssec:ablation_ibp_init_warm_up_loss}
Besides evaluating the use of batch normalisation layers for our Certified
Training scenario, we also check whether the IBP initialisation and warm-up
loss as introduced by \citet{Shi+21a} improve performance for the networks
we train on CIFAR10. To do so, we re-implement both of them and compare the standard
and robust accuracy for networks without their proposed changes, with just
IBP initialisation, with only the warm-up loss and with both IBP
initialisation and warm-up loss added. The networks are trained using the
SSIP loss with the test set robust accuracy computed using SIP bounds as
well. The training runs are repeated for different kernel sizes $s \in \{5,
7\}$ and perturbation strengths $\epsilon \in \{0.2, 0.6, 1.0\}$ while
considering box blur perturbations. The results obtained from these runs
are visualised in Figure \ref{fig:ablation_study_ibp_init_warmup}. It is
easy to see that, irrespective of the perturbation strength and kernel
size, using the IBP initalisation to initialise the network weights or the
warm-up loss during the warm-up phase of the training process does not have
a significant impact on the accuracy of the trained networks. This is true
for both the standard and the robust accuracy. Due to these results, we do
not use IBP initialisation or warm-up losses for our training runs
discussed in Section \ref{sec:evaluation}.
\begin{figure*}[!htb]
\centering
\pgfplotsset{
  every axis plot/.append style={line width=1pt},
  every axis plot post/.append style={
    every mark/.append style={line width=1.6pt}
  }
}
\begin{subfigure}{0.49\textwidth}
\centering
\resizebox{\linewidth}{!}{
\begin{tikzpicture}
	\begin{axis}[
		xmin=0.5,
		xmax=4.5,
		ymin=0,
		ymax=100,
		xlabel=,
		ylabel=Standard Accuracy,
		axis x line=bottom,
		axis y line=left,
		legend style={at={(0.5,-0.2)},    
                     anchor=north,legend columns=4},
		legend to name={mylegend},
		xticklabels={Standard, IBP Init, Warm-Up Loss, {IBP Init + Warm-Up Loss}},
		xtick={1,...,4},
		x tick label style={text width=1.5cm,align=center},
		every y tick label/.append style={font=\small},
		every x tick label/.append style={font=\small},
	]
	\addplot[color=NavyBlue,mark=*] coordinates {
		(1,88.71)
		(2,89.08)
		(3,88.57)
		(4,88.65)
	};
	\addlegendentry{$\epsilon=0.2$}
	\addplot[color=ForestGreen,mark=square*] coordinates {
		(1,84.98)
		(2,84.57)
		(3,84.68)
		(4,84.22)
	};
	\addlegendentry{$\epsilon=0.6$}
	\addplot[color=RedOrange,mark=triangle*] coordinates {
		(1,73.30)
		(2,72.25)
		(3,72.68)
		(4,72.60)
	};
	\addlegendentry{$\epsilon=1.0$}
	\end{axis}
\end{tikzpicture}
}
\caption{Standard Accuracy, kernel size $s=5$}
\vspace{0cm}
\end{subfigure}
\begin{subfigure}{0.49\textwidth}
\centering
\resizebox{\linewidth}{!}{
\begin{tikzpicture}
	\begin{axis}[
		xmin=0.5,
		xmax=4.5,
		ymin=0,
		ymax=100,
		xlabel=,
		ylabel=Robust Accuracy,
		axis x line=bottom,
		axis y line=left,
		legend style={at={(0.5,-0.2)},    
                     anchor=north,legend columns=4},
		legend to name={mylegend},
		xticklabels={Standard, IBP Init, Warm-Up Loss, {IBP Init + Warm-Up Loss}},
		xtick={1,...,4},
		x tick label style={text width=1.5cm,align=center},
		every y tick label/.append style={font=\small},
		every x tick label/.append style={font=\small},
	]
	\addplot[color=NavyBlue,mark=*] coordinates {
		(1,82.67)
		(2,83.09)
		(3,82.71)
		(4,83.39)
	};
	\addlegendentry{$\epsilon=0.2$}
	\addplot[color=ForestGreen,mark=square*] coordinates {
		(1,72.43)
		(2,71.43)
		(3,72.24)
		(4,71.96)
	};
	\addlegendentry{$\epsilon=0.6$}
	\addplot[color=RedOrange,mark=triangle*] coordinates {
		(1,57.70)
		(2,57.57)
		(3,58.63)
		(4,58.11)
	};
	\addlegendentry{$\epsilon=1.0$}
	\end{axis}
\end{tikzpicture}
}
\caption{Robust Accuracy, kernel size $s=5$}
\vspace{0cm}
\end{subfigure}
\begin{subfigure}{0.49\textwidth}
\centering
\resizebox{\linewidth}{!}{
\begin{tikzpicture}
	\begin{axis}[
		xmin=0.5,
		xmax=4.5,
		ymin=0,
		ymax=100,
		xlabel=,
		ylabel=Standard Accuracy,
		axis x line=bottom,
		axis y line=left,
		legend style={at={(0.5,-0.2)},    
                     anchor=north,legend columns=4},
		legend to name={mylegend},
		xticklabels={Standard, IBP Init, Warm-Up Loss, {IBP Init + Warm-Up Loss}},
		xtick={1,...,4},
		x tick label style={text width=1.5cm,align=center},
		every y tick label/.append style={font=\small},
		every x tick label/.append style={font=\small},
	]
	\addplot[color=NavyBlue,mark=*] coordinates {
		(1,88.32)
		(2,88.71)
		(3,88.24)
		(4,88.13)
	};
	\addlegendentry{$\epsilon=0.2$}
	\addplot[color=ForestGreen,mark=square*] coordinates {
		(1,84.21)
		(2,84.34)
		(3,83.29)
		(4,84.20)
	};
	\addlegendentry{$\epsilon=0.6$}
	\addplot[color=RedOrange,mark=triangle*] coordinates {
		(1,68.57)
		(2,68.76)
		(3,68.71)
		(4,67.70)
	};
	\addlegendentry{$\epsilon=1.0$}
	\end{axis}
\end{tikzpicture}
}
\caption{Standard Accuracy, kernel size $s=7$}
\vspace{0cm}
\end{subfigure}
\begin{subfigure}{0.49\textwidth}
\centering
\resizebox{\linewidth}{!}{
\begin{tikzpicture}
	\begin{axis}[
		xmin=0.5,
		xmax=4.5,
		ymin=0,
		ymax=100,
		xlabel=,
		ylabel=Robust Accuracy,
		axis x line=bottom,
		axis y line=left,
		legend style={at={(0.5,-0.2)},    
                     anchor=north,legend columns=4},
		legend to name={mylegend},
		xticklabels={Standard, IBP Init, Warm-Up Loss, {IBP Init + Warm-Up Loss}},
		xtick={1,...,4},
		x tick label style={text width=1.5cm,align=center},
		every y tick label/.append style={font=\small},
		every x tick label/.append style={font=\small},
	]
	\addplot[color=NavyBlue,mark=*] coordinates {
		(1,82.11)
		(2,82.56)
		(3,82.77)
		(4,82.51)
	};
	\addlegendentry{$\epsilon=0.2$}
	\addplot[color=ForestGreen,mark=square*] coordinates {
		(1,70.94)
		(2,70.59)
		(3,70.28)
		(4,70.84)
	};
	\addlegendentry{$\epsilon=0.6$}
	\addplot[color=RedOrange,mark=triangle*] coordinates {
		(1,51.06)
		(2,50.95)
		(3,52.91)
		(4,51.36)
	};
	\addlegendentry{$\epsilon=1.0$}
	\end{axis}
\end{tikzpicture}
}
\caption{Robust Accuracy, kernel size $s=7$}
\vspace{0cm}
\end{subfigure}

\ref*{mylegend}
\caption{Ablation Study on using IBP initialisation and warm-up loss for Certified Training on convolutional perturbations.}
\label{fig:ablation_study_ibp_init_warmup}
\end{figure*}
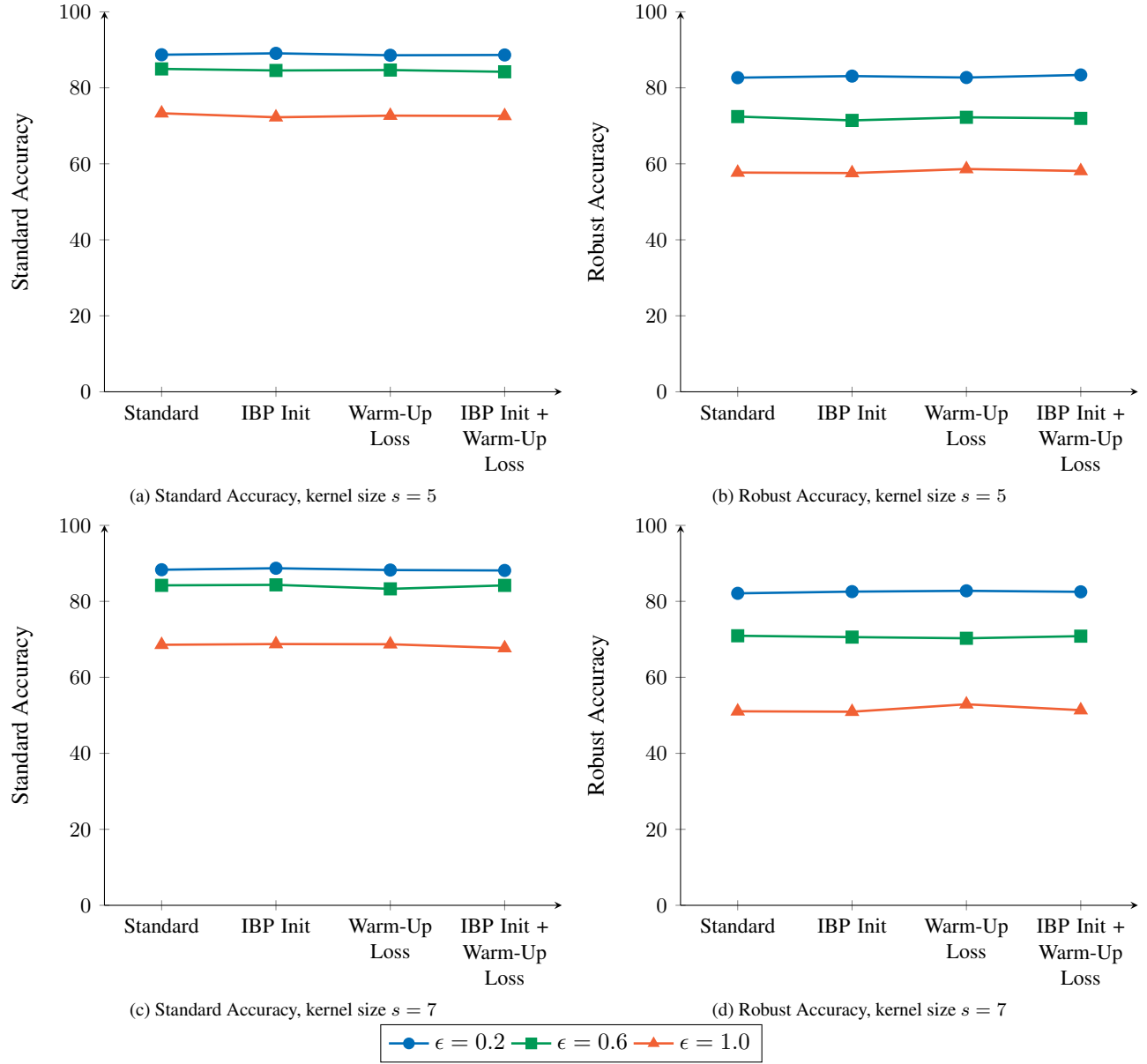

\end{document}